\documentclass{llncs}
\usepackage[rightcaption]{sidecap}
\usepackage{cite}
\usepackage{amsmath,amsfonts}
\usepackage{array,booktabs}
\usepackage{url}
\usepackage[utf8]{inputenc}
\usepackage{graphicx}
\usepackage{authblk}
\usepackage{bm}
\usepackage{enumitem} 
\usepackage{multirow}

\providecommand{\OO}[1]{\mathcal{O}\bigl(#1\bigr)}
\providecommand{\oo}[1]{o\bigl(#1\bigr)}
\providecommand{\B}[1]{\bm{#1}}
\def\slantfrac#1#2{\kern.1em^{#1}\kern-.1em/\kern-.1em_{#2}} 

\begin{document}

\newif\ifanonimize


\mainmatter              
\title{Improving accuracy and power with transfer learning using a meta-analytic database}
\titlerunning{Meta-analysis by transfer-learning}
%
\ifanonimize
\else
\author{Yannick Schwartz\inst{1,2}
  \and Ga\"el Varoquaux\inst{1,2}
  \and Christophe Pallier\inst{3,2}
  \and Philippe Pinel\inst{3,2}
  \and Jean-Baptiste Poline\inst{2}
  \and Bertrand Thirion\inst{1,2}}

\authorrunning{Y. Schwartz et al.} 
%
%
\institute{Parietal Team, INRIA Saclay-\^{I}le-de-France, Saclay, France\\
\email{yannick.schwartz@inria.fr},
\and
CEA, DSV, I\textsuperscript{2}BM, Neurospin b\^{a}t 145,
91191 Gif-Sur-Yvette, France
\and
INSERM, CEA, Cognitive Neuroimaging Unit, Neurospin center, France
}
\fi

\maketitle              

\begin{abstract}
Typical cohorts in brain imaging studies are not large enough 
for systematic testing of all the information contained in the images.
To build testable working hypotheses,
investigators thus rely on analysis of previous work, sometimes formalized in a
so-called \emph{meta-analysis}. 
In brain imaging, this approach underlies the specification of regions of
interest (ROIs) that are usually selected on the basis of the coordinates
of previously detected effects. In this paper, we propose to use a
database of images, rather than coordinates, and frame the problem as
\emph{transfer learning}: learning a discriminant model on a 
reference task to apply it to a different but related new task.
To facilitate statistical analysis of small cohorts, we use a sparse
discriminant model that selects predictive voxels on the reference task
and thus provides a principled procedure to define ROIs. The benefits
of our approach are twofold. First it uses the reference database for
prediction, \emph{i.e.} to provide potential biomarkers in a clinical setting.
Second it increases statistical power on the new task. We demonstrate on
a set of 18 pairs of functional MRI experimental conditions that our
approach gives good prediction. In addition, on a specific transfer
situation involving different scanners at different locations, we show
that voxel selection based on transfer learning leads to higher detection
power on small cohorts.

\keywords{Meta-analysis, fMRI, multiple comparison, machine learning}
\end{abstract}

\section{Introduction}

Multi-subject or multi-condition experiments are the workhorse of
bio-medical imaging research, whether it be drug development or basic 
research. Imaging provides a wealth
of information on the biomedical problem at hand. However the typical
sample size is too small to fully exploit this information. For
this reason, investigators often turn to previous studies in order to formulate 
hypotheses and restrict the search space, \emph{i.e.} select a subset of 
anatomical or functional structures of interest to the current study. 
A typical case is that of early-stage clinical trials, for which the
group size is very small, but that are most often based on previous
results concerning the pathology under study.
However, understanding the literature is increasingly difficult
and requires a systematic approach, that takes the form of a
\emph{meta-analysis}, pooling results from multiple experiments
that address a set of related research hypotheses \cite{Sutton2000}.

In particular, brain imaging studies heavily rely on such meta-analyses
\cite{Wager2009}, 
as the brain is still an ill-understood and complex organ.
In functional Magnetic Resonance Imaging (fMRI) studies, typical group
sizes range from 10 to 20 subjects, which
is not always enough to warrant the reliability of brain-wide analysis\cite{Thirion2007}.
More importantly, the time that can be spent in the scanner by subjects is limited, and
not all interesting experimental conditions will be acquired.%
For this reason, it is common practice to reduce the
study to a set of regions of interest (ROIs) extracted from the 
literature. 
Investigators define these ROIs by extracting locations of peak activations 
from the literature \cite{Yarkoni2011}, or from coordinate databases such as 
BrainMap \cite{Laird2005b}. 
While most of these meta-analyses are conducted on activation coordinates, the
increase of data sharing opens the door to meta-analysis on full brain 
images which results in higher statistical power  \cite{Salimi-Khorshidi2009}.
Previous statistical and modeling work on meta-analysis for fMRI has
focused on better modeling of the reference database
\cite{Wager2009}.

In this work, we are interested in the generalization power of meta-analyses 
on new data.
We introduce a new meta-analysis method using a reference
database of images to guide statistical analysis of a new dataset. In
particular we rely on predictive models, useful to learn biomarkers,
and use them to select relevant voxels in order to increase the statistical
power of a new study.

\section{Methods}

\paragraph{Problem setting}
We start from a reference database made of $l$ experiments, each 
comprising $n^l$ contrasts possibly acquired over multiple subjects.
We denote the brain images by $\B{X}^{l} \in \mathbb{R}^{n^l\!,p}$ with
associated experimental condition $\B{y}^{l} \in \mathbb{R}^{n^l}$. Given a
new experiment, denoted \emph{target}, $(\B{X}^{\star}, \B{y}^{\star}) \in
(\mathbb{R}^{n^\star\!,p}, \mathbb{R}^{n^\star})$, we are interested in
two problems: i) \emph{(biomarkers)} can we predict $\B{y}^{\star}$ from
$\B{X}^{\star}$? ii) \emph{(inference)} can we test hypotheses on the 
links between
$\B{y}^{\star}$ and $\B{X}^{\star}$, for instance in a linear model?
These are ill-posed problems from the statistics standpoint, as
$n^\star \ll p$. The root of the problem is the dimensionality of the
data: medical images are composed of many voxels, typically $p \approx 50\,000$ with fMRI.
This large number of descriptors 
limits statistical inference power due to multiple testing; a problem
that appears in predictive approaches as the curse of dimensionality.
Here, we use our reference database to better condition this statistical problem.

\paragraph{Transfer learning}
The gist of our approach is to learn on some experiments of our
database $(\B{X}^{l}, \B{y}^{l})$ discriminative models that
contain predictive information for the target experiment 
$(\B{X}^{\star}, \B{y}^{\star})$. In machine learning, this problem is
known as \emph{transfer learning} \cite{Pan2010}. The underlying
assumption of transfer learning is the same as that for meta-analysis:
the reference database should contain some common information with the
target experiment. Here we use a simple form of transfer learning:
we train a linear classifier on an experiment in the database that is 
similar from the neuroscientific point of view to the new
data, and use it to predict the labels of the new data.

\paragraph{Selecting predictive features}
We use a sparse linear classifier, specifically an $\ell_1$-penalized
logistic regression. The motivation behind this choice of classifier is
that it produces a sparse set of weights that can be used to select
relevant voxels. In particular, under certain conditions, the classifier
can recover with high probability the complete set of $k$ features in $\B{X}$ that
are predictive of $\B{y}$ for a sample size of $n_\text{min} = \OO{k \log
p}$ \cite{Bach2010}. The logarithmic dependence in $p$ is an appealing 
property in view of the dimensionality of medical imaging datasets.

In practical situations, it can be hard to control the errors on this
feature selection, in particular as it depends on the choice of the
amount of $\ell_1$ penalty. For this reason, 
Meinshausen and B{\"u}hlmann \cite{Meinshausen2010}
introduce randomized variants of sparse estimators, that can be seen as
sampling the posterior probability of selection and keeping only features
that are selected frequently. In particular, they establish
non-asymptotic recovery results for the \emph{randomized lasso}, which
consists in applying the Lasso on random subsamples of the data and
rescaling of the regressors. Here, we adapt this strategy to
classification as the logistic
regression is locally equivalent to a weighted least square and recovery
results can carry from square-loss regression to logistic regression
\cite{Bach2010}.

We want to use transfer learning to perform screening of the voxels,
\emph{i.e.} eliminate many voxels that are not related to our target experiment.
For this purpose, we need a low probability of rejecting relevant
variables. Each iteration of the sparse logistic regression in
the randomized logistic can select reliably only $k_\text{max} \approx n
/ \log p$ variables.
In the worst case situation, we have $k$ heavily-correlated variables and
one of them is selected at random by the sparse logistic
regression at each iteration. For each of these variables, the probability of selecting it
less than $s$ times during $m$ iterations of the randomized logistic is
given by the cumulative distribution function of a binomial with per
trial success ratio $1/k$. If $s \leq m /k$, by
Hoeffding's inequality, this probability goes to 
zero in $\oo{\exp m }$. In other words, if we impose a threshold $\tau =
s/m$ 
on the selection frequency, we can recover a group of $k$ correlated variables 
for $\tau \leq 1/k $.

\paragraph{Brain parcellations}
Although randomization relaxes the conditions on
recovery, a remaining necessary condition is that the regressors of interest,
\emph{i.e.} the
values $\B{x}_i$ across the subjects of the $k$ predictive voxels, must
be weakly correlated\footnote{Specifically, the condition for recovery
with randomized lasso it is a lower bound on the conditioning of the 
sparse eigenvalues of the design matrix \cite[theorem 2]{Meinshausen2010}
and for sparse logistic regression the corresponding condition is a 
lower bound on the eigenvalues of the regressors of interest's 
covariance matrix \cite[theorem 4]{Bach2010}.}. Because of the large
amount of smoothness present in medical images, in particular in group-level 
fMRI contrasts, these conditions cannot be satisfied. Indeed, values taken by a
voxel are very similar to values taken by its neighbors. In addition, the
numbers of subjects used in fMRI are often below the sample size required
for good recovery. For these reasons we resort to feature agglomeration:
using hierarchical clustering to merge neighboring voxels carrying
similar information into parcels
\cite{Michel2011}. This strategy brings the double benefit of
reducing the problem size, and thus the required sample size, and
mitigating local correlation, at the expense of spatial resolution.

\section{Experiments and Results}

\subsection{FRMI datasets}

We use 3 studies for this meta-analysis. The first study (\emph{E1}) 
\cite{Pinel2007} is 
composed of 322 subjects and was designed to assess the inter-subject
variability in some language, calculation, and sensorimotor tasks. The second study 
(\emph{E2}) is similar to the first one in terms of stimuli, but its data was acquired
on 35 pairs of twin-subjects. The last study (\emph{E3}) \cite{Pallier2011}
characterizes brain regions in charge of the syntactic and the semantic
processing for the language. It was performed with 40 subjects, 20 of which
were stimulated by pseudowords (jabberwocky stimuli) instead of actual meaningful
sentences. All the studies were pre-processed and analyzed with the
standard fMRI analysis software SPM5. The data used for this work are the 
statistical images resulting from the intra-subject analyses across the 3 studies. 
\emph{E1} has 34 contrasts images available, \emph{E2} 56, and \emph{E3} 28. The
raw images were not always acquired on the same scanner. \emph{E1} has data from
a 3T SIEMENS Trio, and a 3T Brucker scanner; \emph{E2} data were acquired on a 1.5T
GE Signa; and \emph{E3} images come from the same 3T SIEMENS Trio.

\subsection{Experimental results for prediction}

Here we are interested in the \emph{prediction} problem: using
transfer learning to discriminate a pair of constrasts with an estimator
trained on two other contrasts. 

We used 4 different approaches to learn
the discriminative models. The first approach relies on the activation 
likelihood estimate (ALE) method \cite{Laird2005a}, as this is a commonly
published method for coordinate-based meta-analyses. We extract the activation positions
from the contrasts maps, and then apply a Gaussian kernel. We use 
the preferred FWHM of 
10mm \cite{Turkeltaub2002}. The other approaches directly use the contrast images.
We name \emph{raw contrasts} the method based on the contrasts voxels
values; \emph{contrast-specific parcels} the method that uses parcels
from the training set: and \emph{meta-analystic parcels} the method that
learns parcels from the full database.
We evaluate on our base of contrasts the ability to do transfer
learning, \emph{i.e} to learn decision rules that carry over from one
situation to another. Since we must make the assumption that the reference
contrasts hold common information with the contrasts of interest, we do 
not try out all the possible combinations, but rather manually select pairs of 
contrasts from a single experiment that form a meaningful classification
task (\emph{e.g.}, computation versus 
reading, or Korean language versus French language). Out of all the 
possible combinations, we select 35 pairs of classification task, and 
subsequently combine them 
into 18 transfer pairs, on which it 
is reasonable to think that the transfer could occur (\emph{e.g.},
computation and reading in visual instructions, 
transfer on computation and reading in auditory instructions).
We first train a linear classifier within 6-fold cross validation test on a first
set of pairs, setting the penalization amount by nested cross-validation, we
call this step \emph{inline learning}. We then re-use the
discriminative model on a different pair of contrasts to perform the 
\emph{transfer learning}. The 3 studies containing language related 
tasks, we can transfer between pairs within an experiment, and across experiments.
Among the 18 selected transfer pairs, we find that 9 can give rise to such
a transfer. Since a transfer is directed, we perform it both ways,
which yields once again 18 transfer pairs to test upon.
The associated prediction scores from the different methods are reported 
in Table \ref{tab:transfers}. The general observation is that ALE yields 
a poorer prediction performance than any other method. This is true both 
for the transfer and inline predictions. We also 
find that brain parcellations scores similar to the raw contrasts images, 
and closer to the inline predictions. 
We find that while the contrast-specific parcels and meta-analytic parcels 
methods do not return the same parcels, they produce very close results. 
We can thus use the full database to learn a single reference
parcellation to perform meta-analysis.

\begin{table}[p]
\begin{tabular}{l@{\footnotesize}|c@{\footnotesize}c@{\footnotesize}|c@{\footnotesize}c@{\footnotesize}|c@{\footnotesize}c@{\footnotesize}|c@{\footnotesize}c@{\footnotesize}}
  \hline
  \multicolumn{1}{c}{\footnotesize Names } & 
  \multicolumn{2}{c}{\footnotesize Peaks } &
  \multicolumn{2}{c}{\footnotesize Contrasts } &
  \multicolumn{2}{c}{\footnotesize Parcels } &
  \multicolumn{2}{c}{\footnotesize Meta parcels } \\
  \hline
  & trans. & in. & trans. & in. & trans. & in. & trans. & in. \\
  \hline

\emph{E1}, comp./sent. $\rightarrow$ \emph{E2}, comp./sent. & 0.75 & 0.85 & 0.88 & 0.97 & 0.83 & 0.96 & 0.83 & 0.96 \\
\emph{E2}, comp./sent. $\rightarrow$ \emph{E1}, comp./sent. & 0.66 & 0.83 & 0.88 & 0.96 & 0.85 & 0.95 & 0.85 & 0.96 \\
\emph{E3}, jabb./French (L) $\rightarrow$ \emph{E3}, jabb./French (S) & 0.46 & 0.48 & 0.65 & 0.67 & 0.62 & 0.60 & 0.67 & 0.62 \\
\emph{E3}, jabb./French (S) $\rightarrow$ \emph{E3}, jabb./French (L) & 0.52 & 0.71 & 0.67 & 0.85 & 0.71 & 0.85 & 0.65 & 0.79 \\
\emph{E3}, jabb./French (L) $\rightarrow$ \emph{E2}, Korean/French & 0.65 & 0.46 & 0.73 & 0.79 & 0.65 & 0.81 & 0.76 & 0.85 \\
\emph{E2}, Korean/French $\rightarrow$ \emph{E3}, jabb./French (L) & 0.73 & 0.81 & 0.79 & 0.85 & 0.75 & 0.81 & 0.75 & 0.75 \\
  \hline
\end{tabular}
\caption{ Prediction scores for inline and transfer learning.
trans.= transfer; in.= inline;
comp.= computation, sent.= sentences (reading), jabb.= jabberwocky;
S= sentence with one word constituents, L= one constituent long sentence.
\label{tab:transfers}
}
\end{table}

\begin{figure*}[p]
	\begin{minipage}[b]{.4\linewidth}
	\caption{Prediction performance relative to the best performing
	  approach: inline prediction with raw 
          contrasts images: the p-values indicate whether the associated methods
          are significantly poorer than the best performing method.}
	\label{fig:overview}
	\bigskip
	\end{minipage}
	\hfill
	\begin{minipage}[b]{.57\linewidth}
	\smash{%
	\includegraphics[width=\linewidth]{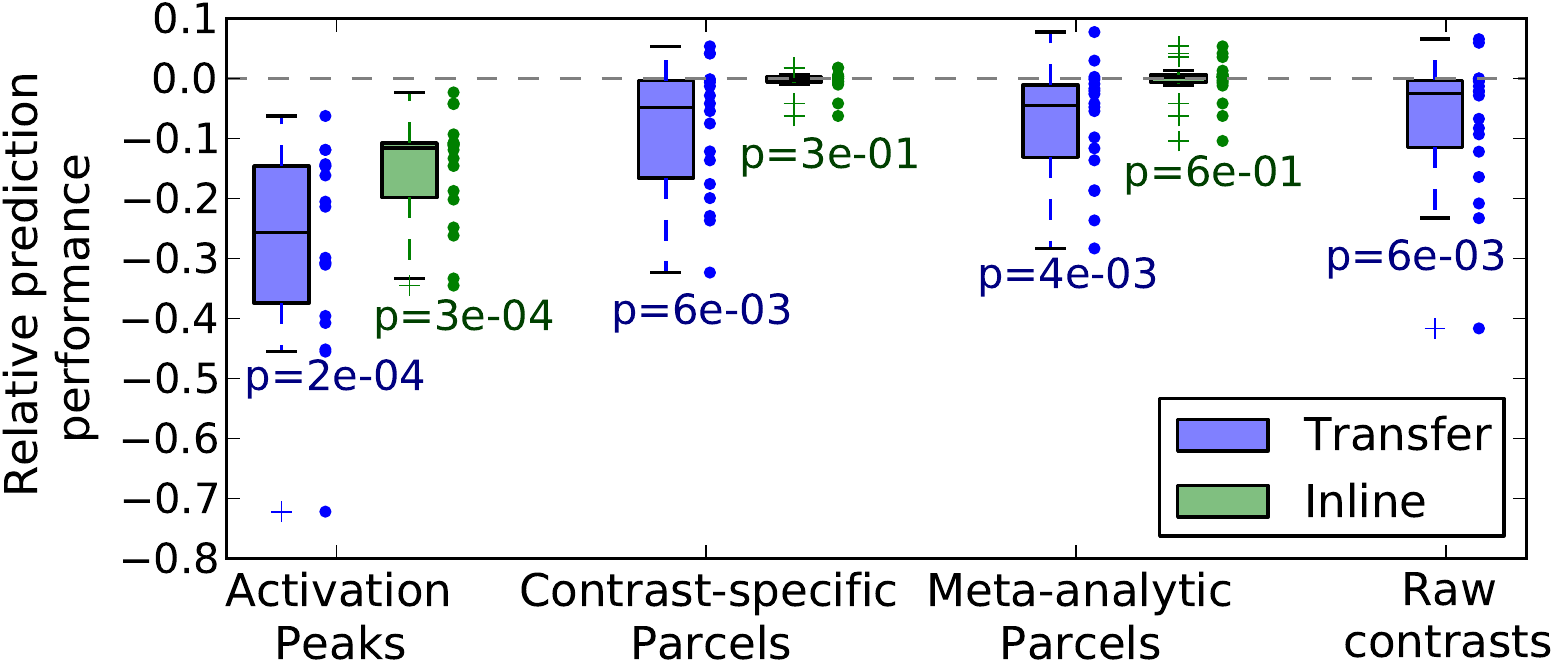}%
	}%
	\end{minipage}%
\end{figure*}

\subsection{Experimental results for inference}

Here we are interested in the \emph{inference} problem: using transfer 
learning to help hypothesis testing on a target dataset.
In the following, we only consider a specific transfer, namely the last
line in Table \ref{tab:transfers}: we learn a model discriminating French
native speakers reading French or Korean, and apply it on another
experiment in which French subjects had to read French or
jabberwocky.
This transfer is interesting as it involves two different
experiments acquired on different scanners, and cognitive paradigms that share a similar
expression, incomprehension of visual language stimuli. As can be seen in 
Table \ref{tab:transfers}, the prediction scores of transfer
learning as well as inline learning on this pair are acceptable although not excellent:
French language and jabberwocky are difficult to separate.

\begin{figure*}[p]
\begin{center}
  \includegraphics{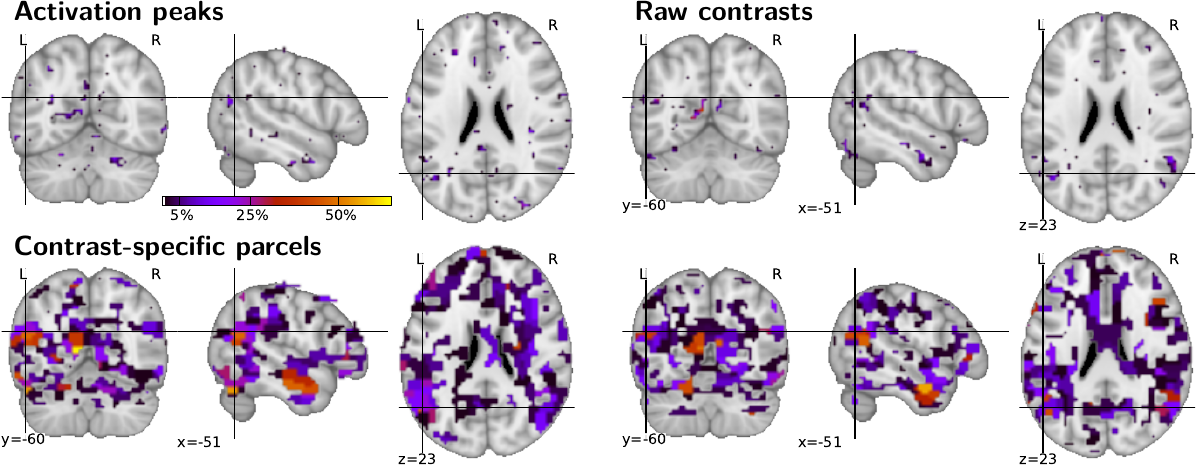}%
  \llap{\raisebox{7.2ex}{\footnotesize\bfseries\sffamily
      \rlap{Meta-analytic parcels}\hspace*{.47\linewidth}}}%

\vskip -0.12in
\caption{Stability scores of the randomized logistic on the Korean versus
French prediction of \emph{E2} for the different set of features: the
colormap represents the frequency at which a feature, parcel or voxel,
was selected. The maps are thresholded at 1\%.}
\label{fig:fmri_results}
\end{center}
\vskip -0.2in
\end{figure*}

Figure \ref{fig:fmri_results} gives the stability scores of the
randomized logistic discriminating reading Korean from reading
French for the different set of features --activation peaks, raw
contrasts, parcels learned on the training contrasts or on the full
database. We can see that while learning at the voxel level or at the
parcel level gives similar prediction performance (Table
\ref{tab:transfers}), the stability score maps are very different. At the
voxel-level, with 70 subjects ($p = 40\,000$, $n = 70$) the recovery is 
limited to
approximately 7 voxels without randomization: the recovery conditions are
violated. As a result, the randomized logistic selects only the most
predictive voxels.
On the parcels, contrast-specific or meta-analytic (i.e., learned on the full
database), the selection frequency highlights regions of the brain that
are known to be relevant for language comprehension, including the left
anterior superior temporal sulcus and the part of the temporal parietal
junction (Wernicke's area). 

We threshold the stability selection scores of the first experiment
(Korean vs French) to select candidate voxels for the target experiment
(jabberwocky vs French). As we want to perform a rough screening and
would rather err on the side of false detections than false rejections,
we take a very low threshold $\tau = .01$. Following our analysis above,
the size of the largest group of correlated features that we can detect 
with such a threshold is on the order of $1 / \tau \approx 100$. With 2000
parcels, this number
corresponds to 5\% of the brain, \emph{i.e.} 8\,000 voxels, and we can
safely consider that no fMRI contrasts is composed of groups of heavily
correlated features larger than this fraction.

\begin{figure*}[p]
\begin{center}
  \includegraphics{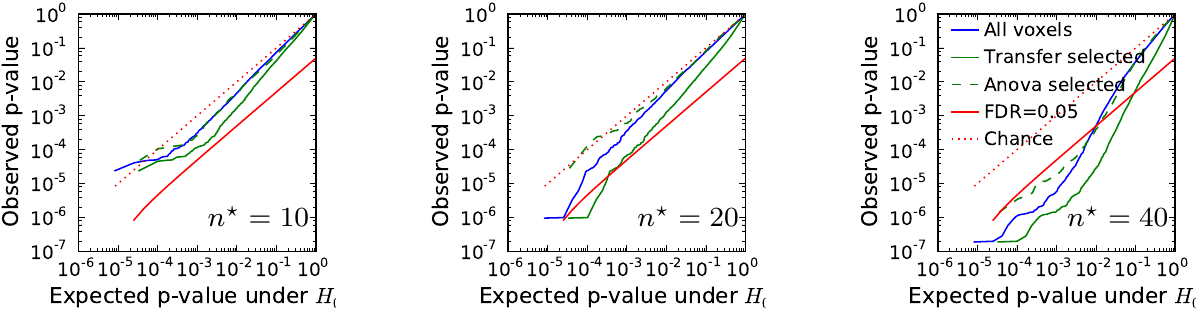}%

\vskip -0.1in
\caption{Q-Q plots for the p-values with and without voxel selection by
transfer learning, as well as FDR=0.05 threshold: {\bf left} for a cohort
size $n^\star = 10$,
{\bf middle} for a cohort size $n^\star = 20$,
{\bf right} for a cohort size $n^\star = 40$.
}
\label{fig:fdr}
\end{center}
\vskip -0.2in
\end{figure*}

On the target experiment, we perform a standard group-level
analysis with the voxels selected, testing for the difference between 
the two conditions,
jabberwocky or French reading. We report results with p-values corrected
for multiple comparisons at a given family-wise error rate (FWER) using
Bonferroni correction, and for a given false discovery rate (FDR) using
the Benjamini-Hochberg procedure.
On table \ref{tab:detections}, we compare
the number of detections and the detection rate, \emph{i.e.} the 
fraction of voxels detected as significantly different, for a full 
brain analysis and for an analysis limited to the voxel selection. We 
compare our voxel selection method to a one-way ANOVA, and find that 
transfer learning outperforms the ANOVA for all the cohort sizes.
Figure \ref{fig:fdr} shows the Q-Q plots on which the Benjamini-Hochberg
procedure is applied. We find that voxel selection by transfer
learning improves both the absolute number of detections and the
detection rate for FWER and FDR correction.

\begin{table}[tb]
\begin{minipage}{1.\linewidth}
\begin{tabular}{r|cc|cc|cc!{\vrule width 1pt}cc|cc|cc}
&\multicolumn{6}{c!{\vrule width 1pt}}{FWER corrected} & \multicolumn{6}{c}{FDR corrected} \\
$n^{\star}$ & \multicolumn{2}{p{1.93cm}|}{All voxels} &
\multicolumn{2}{p{1.6cm}|}{Selection} &
\multicolumn{2}{p{1.6cm}!{\vrule width 1pt}}{ANOVA} &
\multicolumn{2}{p{1.85cm}|}{All voxels} &
\multicolumn{2}{p{1.93cm}}{Selection} &
\multicolumn{2}{p{1.6cm}}{ANOVA} \\
\hline
\hline
$10$ & 0 &(0\%) & 0 &(0\%) & 0 &(0\%) & 0 &(0\%) & 0 &(0\%) & 0 &(0\%)\\
$20$ & 0 &(0\%) & 3 &(0.02\%) & 0 &(0\%) & 0 &(0\%) & 4 &(0.027\%) & 0 &(0\%)\\
$40$ & 5 &(0.0084\%) & 33 &(0.22\%) & 2 & (0.0014\%) & 143 &(0.97\%) & 1339 &(9\%) & 201 & (1.4\%)\\
\hline
\hline
\end{tabular}
\caption{Number of detections at $p < 0.05$ for difference cohort size, for
transfer learning and ANOVA. The percentage of detection is indicated 
in parenthesis.
\label{tab:detections}}

\end{minipage}%

\end{table}

\section{Conclusion}

In this paper, we propose to improve the conditioning and power of statistical 
analyses in imaging studies, using a large meta-analytic database.

In a \emph{transfer learning} scheme, we train on the database
sparse discriminative models that are suited to the target experiment.
Not only can the predictive power of these models can be useful to 
establish biomarkers, but also they perform feature selection 
that can increase the statistical power of a standard
group analysis on new experiments, provided enough predictive features 
(voxels) can be recovered. 
Using brain parcellations, the discriminative model acts to screen
parcels unlikely to be relevant in the target experiment, thus defining 
automatically ROIs.

Using a set of 3 fMRI studies related to language, we confirm experimentally 
that our transfer learning scheme is able to: i) perform accurate predictions on
experiments acquired on a different scanner and with varying paradigm,
ii) outperform the standard meta-analysis procedures based activation peaks,
iii) increase the statistical power in the target experiment by using 
the ROIs defined by the discriminative model.

In this work we manually select the contrast pairs since it is 
delicate to interpret a transfer learning score without good 
knowledge of the cognitive or clinical conditions under study. 
Future work will study automatic contrast pairs selection, 
\emph{e.g.} by mining the descriptions of the experiments 
\cite{Yarkoni2011}, to address 
the problem of synthesizing the ever-growing literature and 
data in medical research.

\paragraph{Acknowledgements}

This work was supported by the ANR grants BrainPedia ANR-10-JCJC 1408-01
and IRMGroup ANR-10-BLAN-0126-02.

\bibliography{biblio}

\end{document}